\documentclass[conference, a4paper]{IEEEtran}
\ifCLASSINFOpdf
\else
\fi

\usepackage{color}

\usepackage{graphicx}
\graphicspath{ {./graphics/} }

\usepackage{multirow}
\usepackage{fancyhdr}
\usepackage{graphicx}
\usepackage{subcaption}
\usepackage{mwe}
\usepackage[nostamp]{draftwatermark}

\begin{document}
%
\title{Predicting Time-to-Failure of Plasma Etching Equipment using Machine Learning}


\pdfoutput=1 

\author{
    \IEEEauthorblockN{Anahid Jalali, \\ 
                      Clemens Heistracher, \\
                      Alexander Schindler \\
                      and Bernhard Haslhofer}
    \IEEEauthorblockA{Austrian Institute of Technology,\\
                      Vienna, Austria\\
                      Email: name.surname@ait.ac.at}

    \and
    \IEEEauthorblockN{Tanja Nemeth, \\Robert Glawar \\ and  Wilfried Sihn}
    \IEEEauthorblockA{Fraunhofer Austria, \\
                      Vienna, Austria\\
                      Email: name.surname@fraunhofer.at}
    \and
    \IEEEauthorblockN{Peter De Boer}
    \IEEEauthorblockA{Infineon Technologies Austria AG,\\
                    Villach, Austria\\
                    Email: Peter.DeBoer@infineon.com}}



\fancypagestyle{firstpage}{%
\renewcommand{\headrulewidth}{0pt}

}

\maketitle

\thispagestyle{firstpage}

\begin{abstract}
Predicting unscheduled breakdowns of plasma etching equipment can reduce maintenance costs and production losses in the semiconductor industry.
However, plasma etching is a complex procedure and it is hard to capture all relevant equipment properties and behaviors in a single physical model.
Machine learning offers an alternative for predicting upcoming machine failures based on relevant data points.
In this paper, we describe three different machine learning tasks that can be used for that purpose: (i) predicting Time-To-Failure (TTF), (ii) predicting health state, and (iii) predicting TTF intervals of an equipment.
Our results show that trained machine learning models can outperform benchmarks resembling human judgments in all three tasks. This suggest that machine learning offers a viable alternative to currently deployed plasma etching equipment maintenance strategies and decision making processes.
\end{abstract}



%

\IEEEpeerreviewmaketitle


\section{Introduction}

Plasma etching is a key procedure in semiconductor wafer fabrication and can, in case of equipment failures or breakdowns, lead to significant production losses. Therefore, maintenance of plasma etching equipment, which aims at minimizing unplanned breakdowns, has become a crucial task. Recently, there has been a clear shift from re-active maintenance planning strategies such as \emph{Run-to-Failure} or \emph{Scheduled Maintenance} to a more pro-active strategy, which is called \emph{Predictive Maintenance (PdM)}. The goal of that strategy is to monitor the health state of an equipment and to predict upcoming failures by estimating the \emph{Time-to-Failure (TTF)} before the next breakdown \cite{matyas2018instandhaltungslogistik}.

Known predictive maintenance approaches can roughly be categorized into \emph{model-based} and \emph{data-driven} methods~\cite{susto2015machine}. Model-based methods rely on domain expertise and knowledge about the physical model of a system in order to predict its degrading behavior. Data-driven approaches, on the other hand, are used when it is not possible to draw a complete picture of a system's physical properties and behaviors. They usually employ machine Learning techniques to model and detect changes in machine behavior. Their effectiveness heavily depends on so called \emph{Health Indicators}~\cite{Guo2017HIRUL}, which are quantitative features that were extracted from available sensor, product quality, and production process data. A selection of relevant features is then used to train a model that describes a machine's health degradation to eventually estimate its remaining lifetime.

A number of studies have already focused on prediction tasks in the plasma etching context: Cheng et al.~\cite{Cheng2003} developed a fault detection and isolation system for plasma etching process chambers. Lou et al.~\cite{luo2015data} used information about a chamber's contamination and employed neural networks for predicting the degradation in a semiconductor manufacturing processes. Puggini and McLoone~\cite{puggini2015} applied \emph{Extreme Learning Machines} to predict the etch rate of each wafer. Munirathinam et al.~\cite{munirathinam2016predictive} predicted a machine's state for maintenance scheduling by focusing on product quality parameters. In summary, existing work focuses on a specific fault type or specific physical properties such as the thickness of the walls caused by particle contamination. 

However, none of those approaches focus on predicting TTF of an entire plasma-etching chamber without directly measuring its health status. Therefore, we studied three different TTF prediction approaches in the context of plasma etching equipment and can summarize our contributions as follows:

\begin{enumerate}

	\item \emph{Task 1}: We modeled TTF prediction as a \emph{regression task} and found that a simple Linear Regression model can be trained to predict the TTF trend and outperform a comparable benchmark resembling human judgment.

	\item \emph{Task 2}: We demonstrate that Task 1 can be transformed into a more effective health state prediction task (\emph{regression}) by converting the TTF target variable.

	\item \emph{Task 3}: We modeled TTF prediction as a \emph{classification task} in order to predict whether breakdowns will occur within defined intervals (e.g., upcoming 0-8h, 8-16h, etc.). We show that standard machine learning algorithms can outperform comparable benchmarks.

\end{enumerate}

All three tasks showed that alarm data, which is defined by the equipment manufacturer, and sensor data limit violations, which are defined by experts operating those equipment, provide the most informative features. We also found that prediction effectiveness is higher in 50-200h intervals than shortly before breakdowns (0-50h).

In the following, we briefly introduce related background information on plasma etching and data-driven predictive maintenance approaches (Section~\ref{sec:background}). Then, in Section~\ref{sec:exmploratory_analysis}, we present key findings of our exploratory analysis before describing our experimental setup in Section~\ref{sec:experiments} and our results in Section~\ref{sec:results}.

\section{Background}\label{sec:background}

\subsection{Plasma Etching}

Semiconductor device production aims to create structures with specific material properties on the surface of a silicon wafer. This can be achieved by selectively adding or removing material and locally changing the chemical structure of the wafer material (e.g., doting and oxidation).

Today, plasma etching is widely chosen for material removal as it provides the high precision level required for the efficiency of modern devices. Prior to the etching stage, areas of the wafer that should not be etched are masked. Next, the wafer is placed in a low-pressure chamber where a plasma is ignited. Ions in this plasma are accelerated towards the wafer and as they hit the surface, they either remove material by mechanical impact, or they form a chemical reaction with the material. The ions in the plasma must be chosen to mostly react with the substrate and not the mask. Additionally, reaction products should be volatile. Otherwise, they might cause deposits on the wafer. To ensure a reliable etching process, control pressures, gas temperatures, gas compositions, voltages, wafer cooling and the plasma composition must be controlled.
 
\begin{figure}[ht]
    \centering
    \includegraphics[width=\columnwidth]{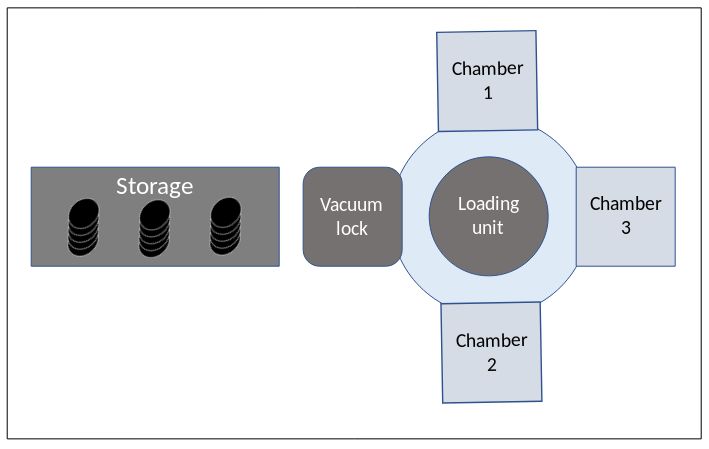}
    \caption{\small{Schematic illustration of chambers and loading mechanism of a modern plasma etching equipment}}
    \label{fig:Scheme_equipment}
\end{figure}

Modern plasma etching equipment operate fully automated and process batches of wafers with predefined recipes, where each recipe is a sequence of predefined process parameters. In general, each equipment has a loading mechanism with vacuum locks and several process chambers. Figure~\ref{fig:Scheme_equipment} illustrates the central functional components of a modern multi-chamber plasma etching equipment.
 
\subsection{Data-Driven Predictive Maintenance Approaches}

Most data-driven predictive maintenance approaches in the literature are based on statistical, probabilistic or machine learning methods and can be categorized into two groups~\cite{Si2013Wienerprocess}: the first group are prognostic models that directly observe production state processes. They either apply regression-based models~\cite{Barraza2017ARX},
or Markovian-based models~\cite{Xiang2012Markovian}. Furthermore, Wang et. al.~\cite{Wang2012HealthPrognostics} proposed a generic model for probabilistic health condition estimation and tested it on two scenarios: (i) electric cooling fans and (ii) an engine dynamic simulation.

Machine learning algorithms such as Support Vector Machines (SVM)~\cite{Patil2015SVMRUL}, Support Vector Regression (SVR)~\cite{Ding2014BatteryRUL} and binary logistic regression~\cite{Phillips2015MachineryConditionclassification} form the second group and have also proved to be an effective solution for estimating TTF and other machine health descriptor variables. 
Deep Neural Networks (DNNs) have recently shown strong performance on variety of complex applications such as speech recognition \cite{Amodei2016deepSpeech}, image classification \cite{Krizhevsky2017NIPS}
and acoustic sound classification and detection \cite{Dang2018SED}. However, as explainability of trained models is a key requirement in the semiconductor industry, other algorithms are typically preferred.  

\subsection{Prediction Maintenance in the Semiconductor Industry}

Outside the specific context of plasma etching, a number of data-driven predictive maintenance approaches have been proposed for monitoring machine health degradation in the semiconductor industry.

Munirathinam et al.~\cite{munirathinam2016predictive} constructed a decision model for a semiconductor fabrication plant by applying a variety of standard machine learning algorithms (e.g., KNNs, SVM). In order to reduce data dimensionality, they applied reduction approaches such as Principal Component Analysis (PCA), Variable Importance Analysis (VIA) and Chi Square statistics. However, although their models can predict whether a product passes a final quality inspection, they were not designed to predict possible equipment breakdowns. 

Lou et al.~\cite{luo2015data}, proposed a two-step maintenance framework for degradation prediction on a real case study in semiconductor manufacturing industries and achieved 74.1\% accuracy in predicting a machine's health degradation. Their work is divided into three stages: first, they adopted a back-propagation Neural network (BPNN) to forecast the machine's health. Second, as a backup for the failed cases of the first stage, they have employed Restricted Boltzmann Machine (RBM) and Deep Belief Networks (DBN), which have strong inductive learning abilities. Finally, using multiple regression forecasting, they checked the prediction accuracy in both stages. However, the model is specialized to predict only the contamination of the process chamber. Failures unrelated to the contamination are not considered in the model. 

Susto and McLoone~\cite{susto2013} adopted an SVM classifier to predict mechanical faults in semiconductor manufacturing. This type of failure is caused by usage and stress of the equipment parts (e.g., filament breaks). In their work, they classify a machine's run as faulty when the SVM's decision boundary falls below a threshold. They specified this threshold by considering two factors: unexpected breakdowns and unexploited lifetime. Their approach showed robustness to cost changes associated with unexpected breaks and inefficient lifetime of an equipment part. However, they do not discuss breakdowns of plasma etching equipment. Additionally, they focus on specific parts (e.g., breakdown of a tungsten filament) and not on a whole plasma etching process chamber.

\section{Exploratory Analysis}\label{sec:exmploratory_analysis}

Before having focused on prediction model building, we first gathered data from several nearly identically constructed plasma etching equipment and process chambers. Then we computed TTF for each chamber, which represents the target variable for all our prediction tasks and the ground-truth for validating trained models. We also conducted an exploratory analysis of available data points in order to identify possible correlations and to reduce the dimensionality of our data. In order to minimize competitive intelligence risks, all figures in this section are schematic and illustrate our findings without exposing details of the underlying manufacturing process.

\subsection{Dataset Characteristics }

Our dataset encompasses data recorded over a 6-month period and has been drawn from the following sources:

\begin{enumerate}

    \item \emph{Automatic Process Control (APC)}: contains 492 distinct sensor data recordings for a single wafer from the underlying plasma etching process control system. This includes statistics of measurements, for instance relevant gas flows and voltages, as well as process-related information such as the time needed for a single etching step and the applied recipes.
    
    \item \emph{APC Limit Violations}: limits are upper- and lower-limit thresholds, which were defined by domain experts for selected APC process control parameters. Limit violations are categorized by severity (\emph{error} vs. \emph{information}) and can trigger actions ranging from automated equipment shutdowns to sending informational emails to domain experts. In total, our dataset contains recordings of 58 different limit violations.

    \item \emph{APC Alarms}: alarms are defined by the machine manufacturer and are categorized into five classes; \emph{warning}, \emph{information}, \emph{critical}, \emph{errors} and \emph{other}. Alarms can, analogous to limit violations, also trigger a number of possible actions, including equipment shutdowns. In total, our dataset contained recordings of 603 distinct alarms per process chamber.

    \item \emph{Real Time Clock (RTC)}: is a system that records state changes (e.g., standby, productive, breakdown) of plasma etching equipment and their corresponding parts. We retrieved those state changes for all equipment over the entire observation period.
    
    \item \emph{Voltage Dips}: describe the voltage reduction in the power supplies. The occurrence of this feature is very sparse. However, domain experts assume that the voltage dips are problematic and can cause equipment's breakdown.

\end{enumerate}

\subsection{TTF Computation}\label{sec:ttf_computation}

\begin{figure}
    \centering
    \includegraphics[width=\columnwidth]{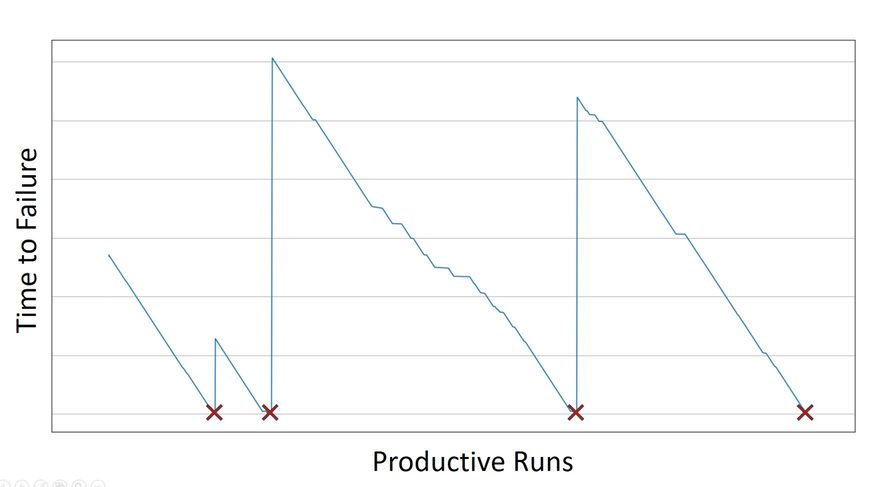}
    \caption{\small{Time-to-Failure (y-axis) of a single chamber over the observation period (x-axis). Red marks indicate recorded breakdowns.}}
    \label{fig:breakdowns}
\end{figure}

In order to build a ground truth for subsequent prediction tasks, we reverse engineered the TTF for each process chamber by joining a machine's state from RTC with process durations extracted from APC. This metric yields a descending counter of productive hours before the next recorded breakdown. Figure~\ref{fig:breakdowns} illustrates the zigzag shape of an arbitrary TTF with four breakdowns.

\subsection{Analysis of APC Limit Violations}

Limit violations can, as discussed before, trigger further actions. An example of a severe violation leading to an immediate shutdown are helium flows above a certain threshold, which are typically caused by dust particles in a chamber. Other violations are only recorded as warnings and do not cause immediate actions.

\begin{figure}
    \centering
    \includegraphics[width=\columnwidth]{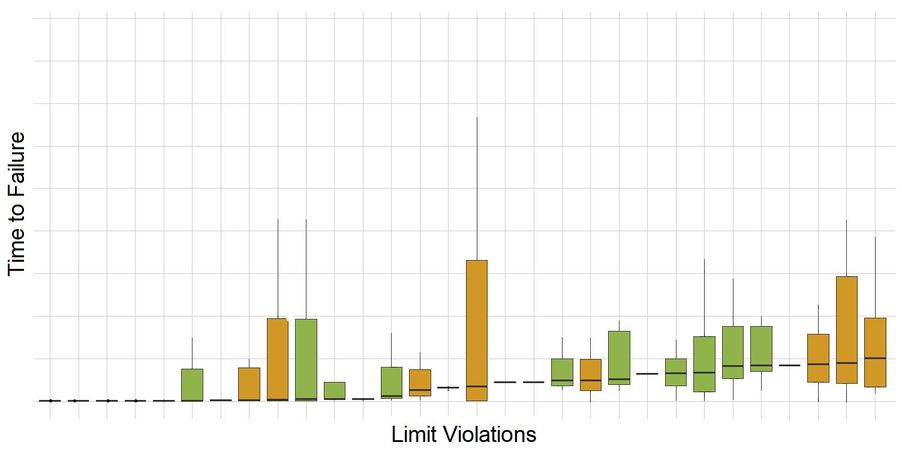}
    \caption{\small{Limit violation occurrences (scatter plot) in relation to TTF. Box plots show medians, colors indicate severity (orange: error; green: information).}}
    \label{fig:violations}
\end{figure}

To understand the impact of limit violations on future process chamber breakdowns, we analyzed historical limit violation recordings and computed the median TTF per defined violation. Figure~\ref{fig:violations} shows a selection of limit violation occurrences (scatter plot) and computed median TTFs (bar plot) in relation to TTF. The single box plots are arranged by their median TTF from lowest to highest, which shows that some violations tend to occur closer to a machine's failure (TTF equals 0) than others. Limit violations with zero TTF are bound to immediate shutdown actions. Intuitively, limit violations that have a median TTF slightly above zero are the most interesting parameters as they indicate upcoming breakdowns but do not immediately cause them. 

\subsection{Analysis of APC Alarms}

\begin{figure}
    \centering
    \includegraphics[width=\columnwidth]{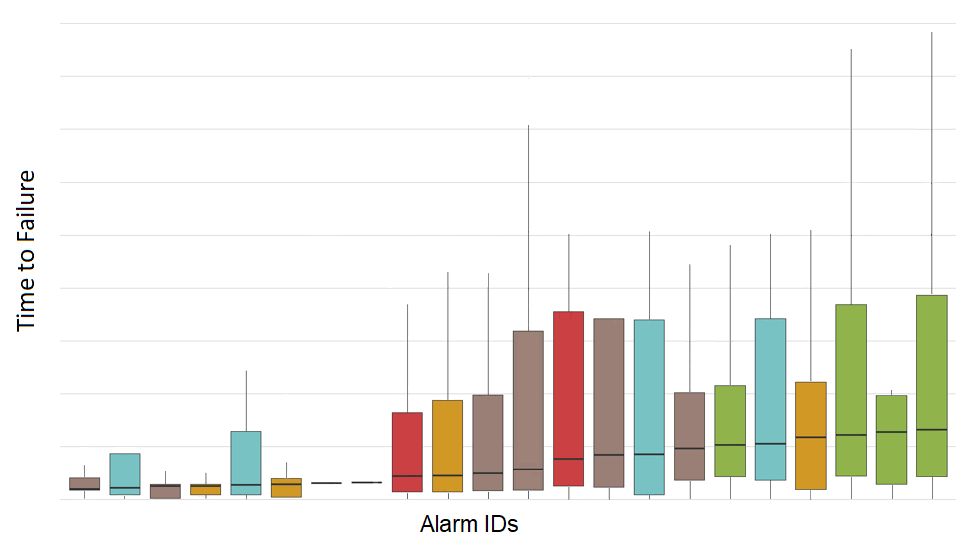}
    \caption{\small{APC alarm occurrences (scatter plot) in relation to TTF. Box plots show medians, colors indicate alarm categories (red: critical; yellow: error; brown: warning; green: information; blue: other).}}
    \label{fig:alarms}
\end{figure}

A plasma etching equipment can raise a number of alarms while processing a wafer. Figure~\ref{fig:alarms} illustrates APC alarm occurrences in relation to a chamber's TTF with each alarm colored by category. Analogous to the previous illustration of limit violations, it arranges alarms in ascending order by their median TTF. Thus, a lower median TTF indicates that an alarm was often raised near an upcoming breakdown. Again, we see that some alarms are more and some are less relevant for our prediction tasks and can assume that the most informative alarms show a median TTF slightly above zero. We also see that alarm severity does not necessarily correspond with the alarm categories defined by the equipment manufacturer.

\subsection{Analysis of APC Sensor Data Correlations}\label{sec:subsec_correlations}

We computed correlations between APC sensor data points in order to identify linear relationships between parameters. This allowed us to reduce the dimensionality of our dataset by 87.2\% by discarding  parameters that correlate strongly with others and therefore add little information. 

\begin{figure}
    \centering
    \includegraphics[width=\columnwidth]{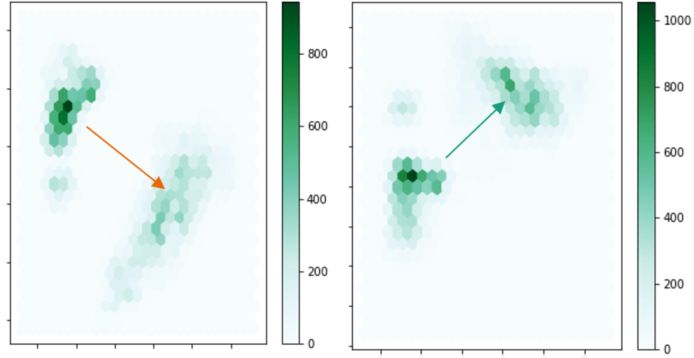}
    \caption{\small{Illustration of APC process variable correlations. Arrows show positive (green) or negative (red) correlations.}}
    \label{fig:cluster-correlation}
\end{figure}

While conducting a principal component analysis, we also found that many process control data recordings strongly depend on the recipe used in an etching process run. This can also be observed when correlating two APC process variables with each other, as shown in Figure~\ref{fig:cluster-correlation}. It shows that correlations are clustered by some external factor, which in this case, is the configured and applied recipe.

\section{Experimental Setup}\label{sec:experiments}

In our experimental setup, we followed the typical data science work-flow: first, we cleansed and normalized our dataset, which had been aggregated from a number of heterogeneous sources. Then, in collaboration with domain experts and based on the observations made in the previous exploratory analysis phase, we engineered a number of features that were potentially useful to the prediction model. Next, in order to evaluate the effectiveness of our models, we built a benchmark resembling human decision making. Finally, we built predictive models for three different TTF prediction and interpretation methods and evaluated their effectiveness against the respective benchmarks. In the upcoming subsections we describe each of those steps in more detail.

\subsection{Data Cleansing and Normalization}\label{sec:experiments_cleansing}

In this step, we removed non-relevant data points such as system-internal identifiers, user names or null-values that do no contribute valuable information to our models.

Given the recipe dependencies of sensor data correlations described in Section~\ref{sec:subsec_correlations}, we standardized all sensor features by the sample mean and standard deviation of the corresponding recipe. This makes the values of each feature in the data have zero-mean and unit-variance and allows comparison across recipes. We also eliminated recipes used by non-productive runs (e.g., cleaning, experiments).

Filtering by relevant recipes reduced the number of missing sensor parameter values to 53\%. We set remaining missing values to zero when we could safely assume that a certain process parameter was not used. Having a NaN in a helium flow, for instance, means that no helium was used in a process step. This also holds for voltage, intensity, and power values.

Finally, we computed the TTF target variable for each chamber, assigned unique identifiers to each segment between breakdowns and normalized that variable to zero mean and unit variance.

\subsection{Feature Engineering}\label{sec:experiments_feature_engineering}

Given the cleansed and normalized datasets, we then engineered a number of feature sets, which fall into three main groups: features derived from APC sensor data (\emph{APC}), alarm features (\emph{AL}), and features derived from limit violations (\emph{LV}).

\begin{figure}
    \centering
    \includegraphics[width=\columnwidth]{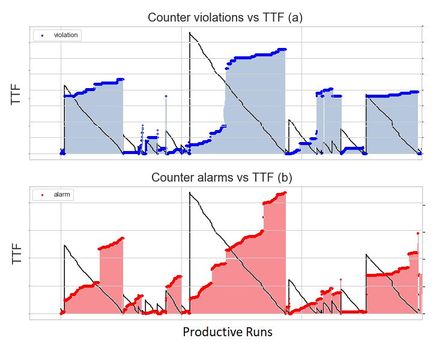}
    \caption{\small{Limit violation and alarm counter features compared to Time-to-Failure of one chamber. Sub-figure (a) depicts the cumulative weighted sum of the limit violations over TTFs of multiple productive intervals. Sub-figure (b) shows the cumulative weighted sum of the alarms within the same interval.}}
    \label{fig:alarm_violation_counter}
\end{figure}

For alarms we constructed a cumulative feature that sums the number of alarm occurrences (\emph{counter alarms}) in each productive interval. Additionally, we weighted each occurrence with a penalty value that captures the severity of an alarm. The penalty for each alarm is computed by inversing its median TTF ($AP_x = 1/median(TTF)_{x}$) and then added to an alarm occurrence. In analogy to alarms, we created a similar feature (\emph{counter violations}) for limit violations. Figure~\ref{fig:alarm_violation_counter} illustrates those weighted alarm and limit violation occurrence features. In sub-figure (a) we can clearly see an increase of violations during the run-time of a chamber and spikes shortly before breakdowns. Sub-figure (b) shows similar behavior, which can be modeled by computing the gradient of those features.

Table~\ref{tb:featuresets} summarizes the feature sets used in our experiments. We selected a number of feature combinations based on feedback from domain experts who defined limit violations and thereby indirectly pointed our importance of a features based on past observations. In the following, we denote, for readability purposes, our feature set combinations as follows: \\[0.5cm]
$FS_{1} = APC_{V} + APC_{R}$\\
$FS_{2} = APC_{V} + LV_{P}$\\
$FS_{3} = APC_{V} + APC_{R} + LV_{P} + AL_{P}$\\
$FS_{4} = APC_{V} + APC_{R} + LV_{P} + AL_{P} +  Voltage Dips$\\
$FS_{5} = APC_{V} + LV_{P} + AL_{P}$\\
$FS_{6} = LV_{P} + AL_{P}$\\
$FS_{7} = AL_{P}$\\


\begin{table}
    \centering
    \caption{\small{Feature sets used in the model}}
    \label{tb:featuresets}
    \begin{tabular}[c]{ |p{1.5cm}|p{6cm}| } 
        \hline
        Feature Set & Description \\[0.01cm]        
        \hline
         $APC_{V}$ & Subset of APC process data containing only features with defined limit violations.  \\
         $APC_{R}$ & Contains information on the recipe mix of x runs (e.g., gradient sum, gradient max). \\ 
         $LV_{P}$ & Engineered cumulative sum of limit violations with penalty (gradient sum, gradient max). \\ 
         $AL_{P}$ & Engineered Cumulative sum of alarms with penalty (gradient sum, gradient max). \\ 
        \hline
    \end{tabular}
\end{table}

\subsection{Model Building}

We have chosen a number of machine learning algorithms that are known (cf. Section~\ref{sec:background}) for their robustness among applications of classification and regression for health monitoring and prognosis. For regression tasks we used Linear Regression (LR), Support Vector Machines (SVM), Decision Trees (DT), Random Forest (RF), and Multilayer Perceptrons (MLP). For the classification task, we used the same algorithms except LR. Additionally, we also chose Gradient Boosting Classifier (GBC), SVM with Stochastic Gradient Descent (SGD) and K-Nearest Neighbours (KNN). For all our models we used implementations of Scikit-learn \cite{pedregosa2011scikit}.

The dataset is split in a way that ensures that recordings from the same productive interval (the time between two breakdowns) are always assigned to the same fold. This is necessary because data in a productive interval is autocorrelated and would reveal information on the next breakdown, if it was used for both training and testing.

To ensure the reliability of our results, we applied a 4-fold cross validation and iteratively trained our models on three folds of data and tested it on one other fold not included in the training data set.

\subsection{Benchmark Definition}\label{Benchmarks}

\begin{figure}
    \centering
    \includegraphics[width=\columnwidth]{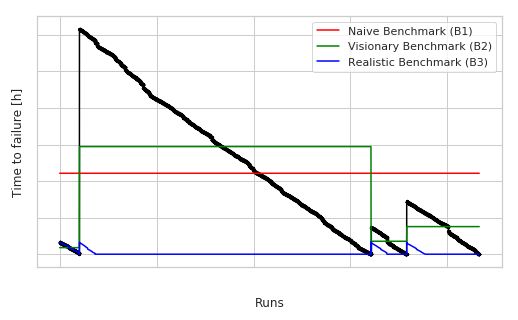}
    \caption{\small{Illustration of three benchmarks used for model evaluation. The red line represents the first benchmark (mean TTFs over entire observation period); the green line the second benchmark (adjusted mean TTF for each productive interval) and the blue line represents the third benchmark (based on historic means).}}
    \label{fig:benchmarks}
\end{figure}

We evaluated the effectiveness of our models in comparison to three benchmarks resembling human decision making: for the \emph{na{\"i}ve benchmark} (B1), we assume that the TTF is always constant and simply take the value of the mean TTF over all runs. The \emph{visionary benchmark (B2)} serves as an idealized reference and considers an adjusted mean TTF for each productive segment, which is unrealistic because it requires knowledge on future breakdowns. The \emph{realistic benchmark (B3)} is closer to reality and assumes that a breakdown occurs after $x$ productive hours where $x$ denotes mean productive time between breakdowns based on historic data. Figure~\ref{fig:benchmarks} visualizes all three benchmarks.

\subsection{Model Evaluation}

We evaluate our models using standard metrics used in machine learning. For regression tasks we compute the root mean squared error (RMSE) of the predicted and the actual value. Then we compute the relative difference to our realistic benchmark B3 as follows: $ x_{rel} = (x -B3)/B3 $ with $x$ denoting the RMSE of the prediction and $B3$ denoting the RMSE of B3. Thus, a negative $x$ characterizes an improvement to the realistic benchmark, whereas positive values denote a prediction that is less effective than the benchmark.

Although the RSME is suitable metric for finding the prediction that is on average closest to the actual TTF curve, it does not necessarily evaluate a model's practical applicability. A constant prediction of the mean value, for instance, might result in a sound RMSE, but is less useful for maintenance purposes as no degradations resembling the decay of TTF is shown. Therefore, we define the best useful model as the model having the lowest RMSE and showing degradation upon visual inspection of predictions.

Furthermore, we evaluate the effectiveness of the classification task with the standard precision (P), recall (R) and F1-score (a trade off between precision and recall). For a more detailed explanation of machine learning evaluation metrics, we refer the reader to related literature such as Powers \cite{powers2011evaluation}.


\section{Results}\label{sec:results}
In the following, we present the results for three types of models we built for predicting the TTF for plasma etching equipment.

\subsection{TTF Prediction (Regression)}

The main goal of this task is to predict the remaining time to a machine's failure. In Section~\ref{sec:ttf_computation} we explained the calculation of the zigzag shaped TTF curve, which is the target variable of this task. To find the most informative set (or sets of features), we used different feature set combinations as input of our regression models with the objective of minimizing RMSE. Afterwards, we compared the results of each experiment with our previously defined benchmarks.

\begin{table}
\center
\caption{\small{Results TTF prediction with \(B3_{RMSE} = 223.96.\)}}
\label{tb:chamber-part-based-RMSE}
\resizebox{\columnwidth}{!}{
\begin{tabular}{|c|c|c|c|c|c|c|c|c|}
\hline
Features & B1                     & B2                     & B3                 & LR             & SVM            & MLP            & Tree           & RF             \\
\hline
FS1      & \multirow{7}{*}{-0.29} & \multirow{7}{*}{-0.49} & \multirow{7}{*}{0} & -0.36 & -0.40           & -0.34          & -0.42 & -0.33 \\
FS2      &                        &                        &                    & -0.31          & -0.40           & -0.3           & -0.18          & -0.29          \\
FS3      &                        &                        &                    & -0.26          & -0.40           & -0.27          & -0.08          & -0.31          \\
FS4      &                        &                        &                    & -0.24          & -0.40           & -0.35 & -0.07          & -0.31          \\
FS5      &                        &                        &                    & -0.26          & -0.40           & -0.29          & -0.05          & -0.31          \\
FS6      &                        &                        &                    & -0.28          & -0.42            & -0.31          & -0.08          & -0.28          \\
FS7      &                        &                        &                    & -0.28          & -0.42            & -0.28          & -0.05          & -0.28         \\
\hline
\end{tabular}
}
\end{table}
Table~\ref{tb:chamber-part-based-RMSE} presents the result of our experiments for one plasma etching chamber after cleaning erroneous breakdown recordings. It shows that Support Vector Machines (SVM) outperformed other models. However, when inspecting predictions visually, we observed that Li\textbf{}near Regression (LR) had the lowest RMSE while showing a degradation in its prediction. Trained with FS1 it outperformed B3 by 36\%, showed 7\% improvement over B1, and was 13\% worse than the second, visionary benchmark B2.

In summary, our first experiments on a single process chamber showed that a trained regression model can predict the degrading trend of a TTF curve, however, with relatively high RMSE. When analyzing the errors, we found that some sequences of relatively short consecutive breakdowns were, according to domain experts' opinions, erroneous recordings caused by equipment starts and almost immediate shutdowns within maintenance operations.

\subsection{Health State Prediction (Regression)}

\begin{figure}
    \centering
    \includegraphics[width=\columnwidth]{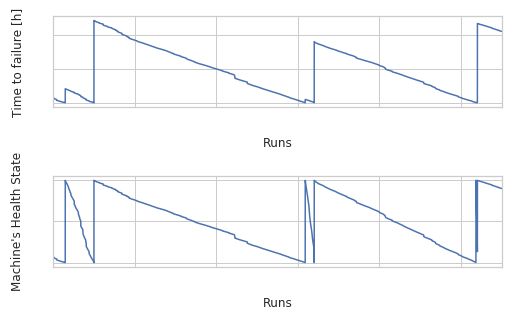}
    \caption{\small{Illustration of time to failures in hours and the corresponding machine's health state.}}
    \label{fig:scaled_TTF_health_state}
\end{figure}

The goal of this task is to predict a machine's health status. For this purpose, we transformed the target value (TTFs) into a range between 0 and 1 (see Figure~\ref{fig:scaled_TTF_health_state}), where a health state of 1 is considered as healthy and 0 as failure (breakdown). All other pre-processing steps were the same as in the first task. 

\begin{figure}
    \centering
    \includegraphics[width=\columnwidth]{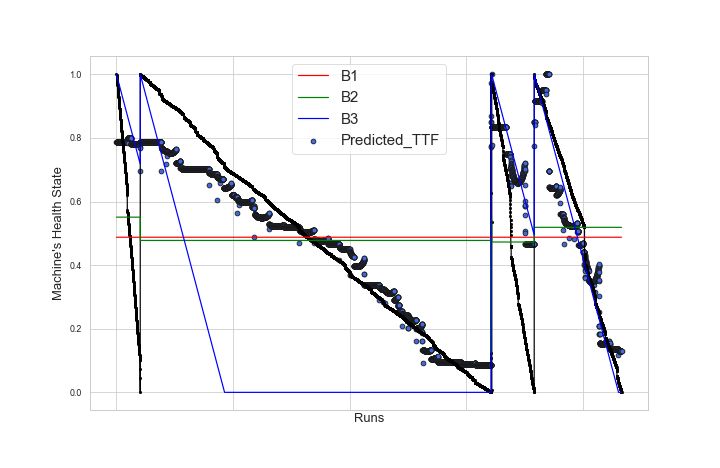}
    \caption{\small{Machine's health state prediction result of a Linear Regression model from one cross-validation fold.}}
    \label{fig:health-state-visualization}
\end{figure}

Table~\ref{tb:chamber-part-based-Health} shows that both Multilayer Perceptron (MLP) and Linear Regression (LR) models trained on all feature set combinations can outperform all benchmarks and visual inspection of predictions indicates degradation in both model families. Figure~\ref{fig:health-state-visualization} illustrates health state prediction results of a trained Linear Regression model from one cross-validation fold.

\begin{table}
\caption{\small{Results of Machine Health prediction  with \(B3_{RMSE} = 0.36.\)}}
\label{tb:chamber-part-based-Health}
\resizebox{\columnwidth}{!}{

\centering
\begin{tabular}{|c|c|c|c|c|c|c|c|c|}

\hline
Features & B1                     & B2                     & B3                 & LR             & SVM            & MLP            & Tree           & RF             \\
\hline
FS1      & \multirow{7}{*}{-0.19} & \multirow{7}{*}{-0.19} & \multirow{7}{*}{0} & -0.22 & -0.19          & -0.22          & -0.03 & -0.19 \\
FS2      &                        &                        &                    & -0.17          & -0.19          & -0.28 & -0.03 & -0.19 \\
FS3      &                        &                        &                    & -0.11          & -0.19          & -0.25          & 0.11           & -0.17          \\
FS4      &                        &                        &                    & -0.11          & -0.19          & -0.25          & 0.11           & -0.19 \\
FS5      &                        &                        &                    & -0.11          & -0.19          & -0.25          & 0.11           & -0.17          \\
FS6      &                        &                        &                    & -0.14          & -0.22 & -0.25          & 0.08           & -0.19 \\
FS7      &                        &                        &                    & -0.14          & -0.22 & -0.25          & 0.06           & -0.19    \\
\hline
\end{tabular}
}
\end{table}

\subsection{TTF Prediction (Classification)}
\label{Task3}

In this task, we consider TTF prediction as a binary classification task that predicts whether a machine will face a breakdown within a predefined time interval of $0$ and $x$, where $x \in \{8h, 16h, 24h, 48h, 72h, 96h, 120h, 144h, 168h, 336h \}$. For a given interval, a run is labeled true when its TTF lies within that interval, otherwise the run is labeled as false.

\begin{figure}
    \centering
    \includegraphics[width=\columnwidth]{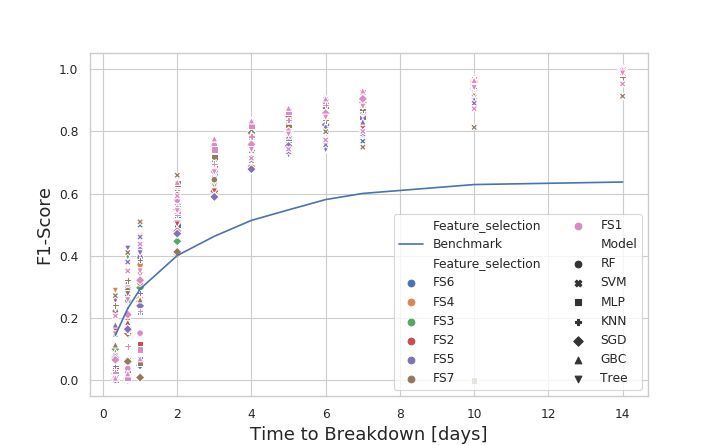}
    \caption{\small{F1-score values in multiple experiments. Each selected feature set is defined with different color and each model is determined with a shape.}}
    \label{fig:fscore-visualization}
\end{figure}

Figure~\ref{fig:fscore-visualization} shows the most effective (highest F1-score) model and feature set combination for each interval. We can clearly observe that trained models can outperform the realistic benchmark (B3) in all intervals. Furthermore, we see that prediction models trained for shorter intervals (e.g., 0-8h) have a lower F1-score than those for longer intervals. This reflects our findings from the previous regression tasks, which also yielded higher RMSE shortly before breakdowns.


A summary of our experimental classification results is presented in Table~\ref{tb:classification-results} which compares the effectiveness of each interval-specific trained model to the corresponding interval-specific benchmark. As in Task 1 and 2 we can observe the importance of alarms and limit violations for shorter predictions as they are included in feature sets FS4, FS5 and FS7 but not in FS1.

\begin{table}[h]
\caption{\small{Summary of best classification results per interval}}
\label{tb:classification-results}

\resizebox{\columnwidth}{!}{
\centering
\begin{tabular}{|c|c|c|c|c|c|c|c|c|}
\hline
\multicolumn{1}{|c|}{\multirow{2}{*}{Interval}} & \multicolumn{3}{c|}{Realistic Benchmark}                                                                           & \multicolumn{5}{c|}{Best Results}                                                                                                                                       \\ \cline{2-9} 
\multicolumn{1}{|c|}{}                          & \multicolumn{1}{c|}{{P}} & \multicolumn{1}{c|}{{R}} & \multicolumn{1}{c|}{{F1}} & \multicolumn{1}{c|}{Features} & \multicolumn{1}{c|}{Model} & \multicolumn{1}{c|}{P} & \multicolumn{1}{c|}{R} & \multicolumn{1}{c|}{F1} \\ \hline
0-8 h                                           & 0.09                               & 0.35                            & 0.15                              & FS4                                    & Tree                       & 0.25                           & 0.36                        & 0.29                          \\
0-16 h                                          & 0.17                               & 0.35                            & 0.23                              & FS5                                    & Tree                       & 0.36                           & 0.53                        & 0.42                          \\
0-24 h                                          & 0.24                               & 0.37                            & 0.29                              & FS7                                    & SVM                        & 0.39                           & 0.75                        & 0.51                          \\
2 days                                          & 0.41                               & 0.39                            & 0.4                               & FS7                                    & SVM                        & 0.59                           & 0.76                        & 0.66                          \\
3 days                                          & 0.54                               & 0.41                            & 0.46                              & FS1                                    & GBC                        & 0.68                           & 0.91                        & 0.78                          \\
4 days                                          & 0.64                               & 0.43                            & 0.51                              & FS1                                    & GBC                        & 0.74                           & 0.96                        & 0.83                          \\
5 days                                          & 0.72                               & 0.44                            & 0.55                              & FS1                                    & GBC                        & 0.79                           & 0.98                        & 0.88                          \\
6 days                                          & 0.8                                & 0.46                            & 0.58                              & FS1                                    & GBC                        & 0.84                           & 0.99                        & 0.91                          \\
7 days                                          & 0.85                               & 0.47                            & 0.6                               & FS1                                    & GBC                        & 0.88                           & 0.99                        & 0.93                          \\
14 days                                         & 1                                  & 0.47                            & 0.64                              & FS6       & MLP                        & 1                              & 1                           & 1                            \\
\hline
\end{tabular}
}
\end{table}


\section{Discussion}

With the overall goal of predicting Time-to-Failure of plasma etching equipment in the semiconductor industry, we experimented with three different machine learning approaches. In all three tasks we were able to outperform comparable benchmarks resembling human breakdown predictions based on historic mean TTF observations. We highlighted the importance of precise breakdown recordings and subsequent TTF computations for building accurate and effective prediction models. We also emphasized the importance of including manufacturer-defined alarms and expert-defined limit violations, which both capture a high degree of domain knowledge. 

One limitation of our approach lies in the relatively high manual effort required for data cleansing, normalization and feature engineering. In order to support genericity and applicability of our overall method on other equipment types with similar data sources, we strongly focused on building generic features that are derived from standard data sources (sensor data, alarms, limit violations) and therefore applicable across equipment. However, we believe that further boosts in effectiveness are possible by modeling equipment-specific behaviors. A possible strategy is to investigate Deep Learning methods, which support automated feature learning but still have the drawback of being non-transparent and hardly explainable. Activation-based or gradient-based methods for feature isolation are possible solutions for those problems.

Finally, we can identify two main orthogonal challenges that need to be tackled when implementing data-driven predictive maintenance in a production setting. First, data quality, especially recordings of breakdowns, is a key prerequisite for building effective prediction models. The most elaborate machine learning method won't provide more precision if the target variable (TTF) is flawed. Second, data-driven breakdown predictions must be embedded into existing maintenance management workflows and operations. This typically requires formation of dedicated groups of professionals having both knowledge of machine and process design as well as an understanding and intrinsic interest in novel data-driven maintenance technologies.


\section{Conclusion}

In this paper, we described three different machine learning tasks that can be used for predicting Time-to-Failure (TTF) or the health state of plasma etching equipment in the semiconductor industry. Our results show that trained prediction models provide acceptable effectiveness for periods exceeding roughly 24 hours, which allows maintenance planners to react on those predictions. We also highlighted the importance of alarms and limit violations, which carry high degrees of domain knowledge. Since model transparency was a key requirement, we restricted ourselves to manual feature engineering and well-known machine learning techniques. In future, however, we will proceed and investigate explainable Deep Learning techniques for those prediction tasks.

\section{Acknowledgement}\label{sec:acknowledgement}
This work was partly funded by the Austrian Research Promotion Agency (FFG) through the project DeepRUL (Project ID: 871357).


\end{document}